\documentclass[final]{cvpr}

\usepackage{times}
\usepackage{epsfig}
\usepackage{graphicx}
\usepackage{amsmath}
\usepackage{amssymb}
\usepackage{bm}
\usepackage[linesnumbered,ruled,vlined]{algorithm2e}
\usepackage{algcompatible}
\usepackage{multirow}
\usepackage{subfigure}
\usepackage{enumitem}

\usepackage[pagebackref=true,breaklinks=true,colorlinks,bookmarks=false]{hyperref}
\def\cvprPaperID{2468} 
\def\confYear{CVPR 2021}

\begin{document}
	
\title{Focus on Local: Detecting Lane Marker from Bottom Up via Key Point}
\author{
    \vspace{0.3em} Zhan Qu\thanks{Z. Qu is the corresponding author.} \enspace Huan Jin \enspace Yang Zhou \enspace Zhen Yang \enspace Wei Zhang\\
    {Noah's Ark Lab, Huawei Technologies}\\
    {\{quzhan, jinhuan3, zhouyang116, yang.zhen, wz.zhang\}@huawei.com}
}


\maketitle
\thispagestyle{empty}

\begin{abstract}
	Mainstream lane marker detection methods are implemented by predicting the overall structure and deriving parametric curves through post-processing. Complex lane line shapes require high-dimensional output of CNNs to model global structures, which further increases the demand for model capacity and training data. In contrast, the locality of a lane marker has finite geometric variations and spatial coverage. We propose a novel lane marker detection solution, FOLOLane, that focuses on modeling local patterns and achieving prediction of global structures in a bottom-up manner. Specifically, the CNN models low-complexity local patterns with two separate heads, the first one predicts the existence of key points, and the second refines the location of key points in the local range and correlates key points of the same lane line. The locality of the task is consistent with the limited FOV of the feature in CNN, which in turn leads to more stable training and better generalization. In addition, an efficiency-oriented decoding algorithm was proposed as well as a greedy one, which achieving 36\% runtime gains at the cost of negligible performance degradation. Both of the two decoders integrated local information into the global geometry of lane markers. In the absence of a complex network architecture design, the proposed method greatly outperforms all existing methods on public datasets while achieving the best state-of-the-art results and real-time processing simultaneously.
\end{abstract}

\section{Introduction}
In autonomous driving system \textbf{(\textit{ADS})}, lane detection plays an important role. On the one hand, the location of host and other traffic participants in the lane forms the basis of autonomous driving decisions. On the other hand, the geometry of a lane marker can be viewed as an important \textit{landmark} of the environment and aligned with a high-resolution or vector map for high-precision positioning. At the same time, lane detection has been widely used in Advanced Driver Assistance Systems \textbf{(\textit{ADAS})} and is the basis for some common features such as Lane Keep Assist \textbf{(\textit{LKA})} and Adaptive Cruise Control \textbf{(\textit{ACC})}.

Recent advances in lane detection can be attributed to the development of convolutional neural networks \textbf{(\textit{CNN})}. Most existing methods adopt well-studied frameworks such as semantic segmentation and object detection to parse lane markers and transform the network output into parametric curves through post-processing. However, the mostly used frameworks can not be seamlessly generalized to curved-shaped lane lines because lane detection task requires precise representation of local positions and global shapes simultaneously, showing their own limitations.

\begin{figure}[htbp]
	\centering
	\subfigure[Segmentation based method and intermediate results.]{
		\includegraphics[width=7.5cm]{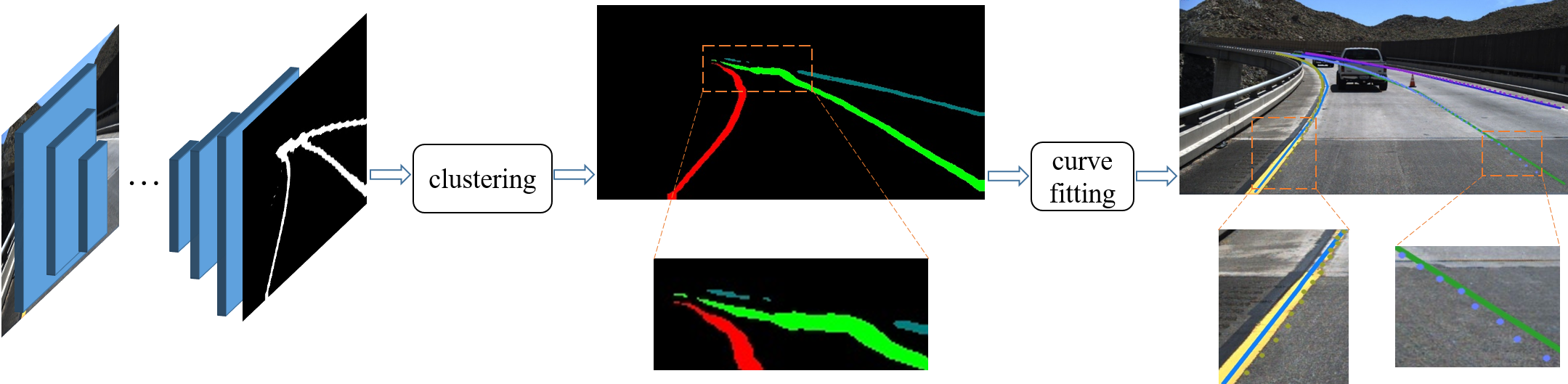}
		\label{fig:seg_fault}
	}
	\quad
	\subfigure[Object detection based method.]{
		\includegraphics[width=7.5cm]{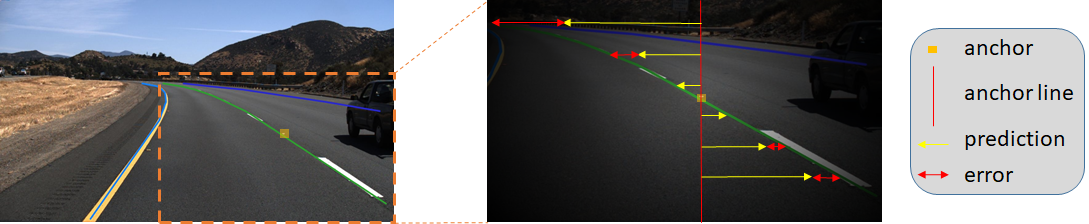}
           \label{fig:reg_fault}
	}
	\caption{Pipeline of existing methods and illustration of common error prediction. \textbf{(a).} the enlarged window in the middle of the pipeline shows the incorrect clustering of the segmentation mask, and the other two orange windows in the curve fitting output show the position deviation of the prediction curve due to redundant and noisy pixels, note the solid line is ground truth, dotted line is prediction. \textbf{(b).} shows the influence of FOV in detection based methods, the brightness of the right image reflects the practical FOV of anchor.}
	\label{examples of disadvantages of previous methods.}
\end{figure}

The semantic segmentation-based approach predicts binary masks of lane marker regions, inserts clustering models into training and inference, groups masked pixels into individual instances, and finally uses curve fitting to parametric results. However, the clustering procedure complicates the training and inference pipeline. In addition, pixel-level inputs to curve fitting are often redundant and noisy, all of which bring negative impact to the accuracy of the final results. Fig.\ref{fig:seg_fault} shows several cases where the prediction errors may increase. Object detection approaches are originally designed for compact target and produce bounding box as output, which is insensitive to pixel-level error when faced with large-scale object. As for lane markers, they typically span half or more of the image, and pixel-level localization errors significantly impair detection performance, which can be attributed to the limited field of view \textbf{(\textit{FOV})} of features learned through CNN being insufficient to model content that is too far apart. Fig.\ref{fig:reg_fault} illustrates the effect of FOV in complex scenario. Moreover, most of these solutions model global geometry directly, and the network must produce high-dimensional outputs to describe the curves. Theoretically, however, uncompact outputs increase the demand for data and model capacity, ultimately masking the generalization ability of the resulting model.

Although the global structure of lane markers has some complexity, we note that local lane markers are extremely simple and that global lane markers can be approximated by a combination of local line segments. Moreover, spatial locality is more suitable for modeling with CNN. Following this intuition, a novel lane marker detection method, FOLOLane, is proposed that focuses on modeling local geometry and integrating them into the global results in a bottom-up manner. Specifically, the geometry of the lane marker is predicted by estimating adjacent \textit{keypoints} on the it. In the bottom stage, a fully convolutional network is used to capture keypoints in the local scope through two separate heads. The first one gives the probability that keypoints appear in pixel space, and the second one gives the offset between keypoints and the most spatially correlated local lane marker, which is used to refine the positions of keypoints generated by the first head and construct associations between keypoints on the same lane markers. Based on the local information, two decoding algorithms with different preferences are proposed to predict global geometry of lane markers. The bottom-up pipeline of the proposed method is shown in Fig.\ref{fig:network}.

Compared with existing works \cite{chen2019pointlanenet, CurveLane-NAS, qin2020ultra, neven2018towards, pan2017spatial}, the proposed approach concentrates the capabilities of CNN on a local scale, which is suitable for CNN's limited FOV, and significantly reduces the complexity of the task and the dimension of the output. As a result, the compact output leads to stable and efficient training without additional effort in network architecture design and data collection. Considering the continuity of lane markers, the proposed decoder is able to associate keypoints of the same instance and optimize the geometry of network predictions without affecting performance and efficiency. Furthermore, during network training and instance decoding, we model and predict keypoints using features with the highest spatial correlation guided by coarse-to-fine strategies. The proposed bottom-up solution achieves the best state-of-the-art level, \textit{\textbf{Acc: 96.92\%}} on TuSimple and \textit{\textbf{F1 score: 78.8\%}} on CULane, and excellent generalization in the two public datasets. Together with the compatibility with network architectures, our approach shows a promising application future.

We emphasize that our method is the first to formulate lane detection into multi-key-points estimation and association problem, which is inspired by the bottom-up human pose estimation framework ~\cite{pishchulin2016deepcut, cao2017realtime, efrat2020semi}. The proposed local scope based method avoids the inaccurate prediction where far from the anchor, which occurs in detection-based methods. And the sparsity of key points prevents the noisy and redundant output occurred in segmentation-based methods, which decrease the precision and increase the delay of curve fitting. With extensive experiments, our solution proves the potential of applying pose estimation approaches on lane detection, which opens up a new direction to solve this important application problem. Our solution does not depend on \textit{CNN} architecture, is readily compatible to newly developed architecture and shows scalable potential on accuracy and efficiency. 

Our contributions can be summarized as follows:
\begin{itemize}
	\item Lane detection is firstly decompose into subtasks of modelling local geometry, which is achieved by estimating keypoints on local curve. Simplified targets and focus on spatially limited scope helps the network to provide precise estimation of local curve.
	
	\item Two decoding algorithms with different preferences are designed to integrate local information into global prediction, which enable the system to achieve high accuracy in ultra real time.
	
	\item Experimental results showed that our approach outperforms all existing methods by a substantial margin. Besides, our model shows the best generalization ability in comparison, which further proves the potential for productization.
\end{itemize}

\thispagestyle{empty}
\section{Related Work}

\textbf{Lane Marker Detection.} Lane marker detection based on deep learning can be categorized into two groups: detection based and segmentation based. The former one: ~\cite{chen2019pointlanenet} proposed an anchor-based lane marker detection model for forward-looking cameras. Lane markers were uniformly sampled along the vertical axis in the image, and dense regression was performed by predicting the offset between each sample point and an anchor line, then  Non-Maximum Suppression\textbf{(\textit{NMS})} was applied to suppress the overlapping detection and select the best lane marker with the highest score. ~\cite{CurveLane-NAS} proposed the use of neural architecture search\textbf{(\textit{NAS})} to find a better backbone and a point blending based post processing to further improve the performance of lane marker detection task. ~\cite{ko2020key} proposed to train a CNN to predict the existence, position and feature embedding of lane markers in an image. A lane marker instance was clustered based on the trained feature embedding. ~\cite{qin2020ultra} formulated lane marker detection as a pixel-wise classification problem for each row of an image. A specific feature map was predicted to indicate the position of a lane marker on each row.

Segmentation based: ~\cite{lee2017vpgnet} proposed a multitask framework, which predicted pixel-wise multi-label and clustered the pixels belonging to same lane instance in bird eye view image using DBSCAN. It also added an auxiliary task: vanish point estimation, to increase the stability of lane marker detection. ~\cite{neven2018towards} proposed an end-to-end joint semantic segmentation and feature embedding network architecture. Pixels on the same lane marker were assigned an identical instance id. ~\cite{pan2017spatial} also designed an instance segmentation network for lane marker detection problem. Different from ~\cite{neven2018towards}, ~\cite{pan2017spatial} predicted a probability map for each lane marker separately and used cubic splines to fit it. In stead of using pixel-wise classification, \cite{yoo2020end} introduced a row-wise classification architecture. For each row, it predicted the most possible grid of a lane marker in an image and recovered a lane marker instance through post processing. ~\cite{liu2020lane} proposed a CycleGAN based method to enhance lane detection performance in low light conditions. ~\cite{ghafoorian2018gan} claimed a more accurate method by using EL-GAN for lane marker detection, which used a generator to segment the lane markers and a discriminator to refine the segmentation result. ~\cite{hou2019learning} proposed a self-attention distillation method for lane marker segmentation task by forcing shallow layers to learn rich context feature from deep layers. 

\textbf{Bottom-Up Human Key Point Detection.} ~\cite{pishchulin2016deepcut} proposed a bottom-up method for crowded scenes, which detected keypoints and built a densely connected graph, the weight of each edge represented the correlation of two keypoints. By optimizing the graph, keypoints belonging to one person were clustered. ~\cite{cao2017realtime} predicted a heat map for each keypoint and part affinity fields (PAFs) which were used to associate body parts with individuals in the image. Similar to ~\cite{efrat2020semi, neven2018towards}, ~\cite{newell2017associative} introduced feature embedding to facilitate keypoints clustering of one person while predicting the heat map of keypoints. ~\cite{papandreou2018personlab} further split the problem into two stages: (1) predicting heat map and short-range offset for keypoints detection, (2) clustering key points using mid-range offset for one person.

We find that lane marker detection can be abstracted as discrete keypoints detection and association problem, which is very similar to bottom-up human key point detection task. ~\cite{philion2019fastdraw} proposed a method based on this idea. A network was trained to extract all possible lane marker pixels and output the pixels in the neighboring row, which belongs to the same lane as the current lane marker pixel. As the problems discussed above, the inherent segmentation-based method inhibited the precise representation of a lane marker. In addition, the pixel-wise joint distribution prediction was redundant.
\thispagestyle{empty}
\section{Methodology}
As shown in Fig.\ref{fig:network}, we proposed a bottom-up lane detection method by estimating the existence and the offsets of the local lane point through the network, followed by a novel global geometry decoder to generate the final curve instances.

\subsection{Network for local geometry}
In the proposed approach, each predicted marker curve is represented as an ordered keypoints set, where the key points are of fixed/predefined vertical interval $\Delta y$ across neighboring rows. First of all, the task of curve prediction is decomposed into local subtasks via a fully convolutional network with two heads. The heatmap outputted by the first head expresses the possibility that keypoint appears, which resolves the existence of local curves. The second head predicts offsets to key points of the most closed local curve, which describes the precise geometry of the local curve.

\begin{figure}[t]
	\begin{center}
		\includegraphics[width=.9\linewidth]{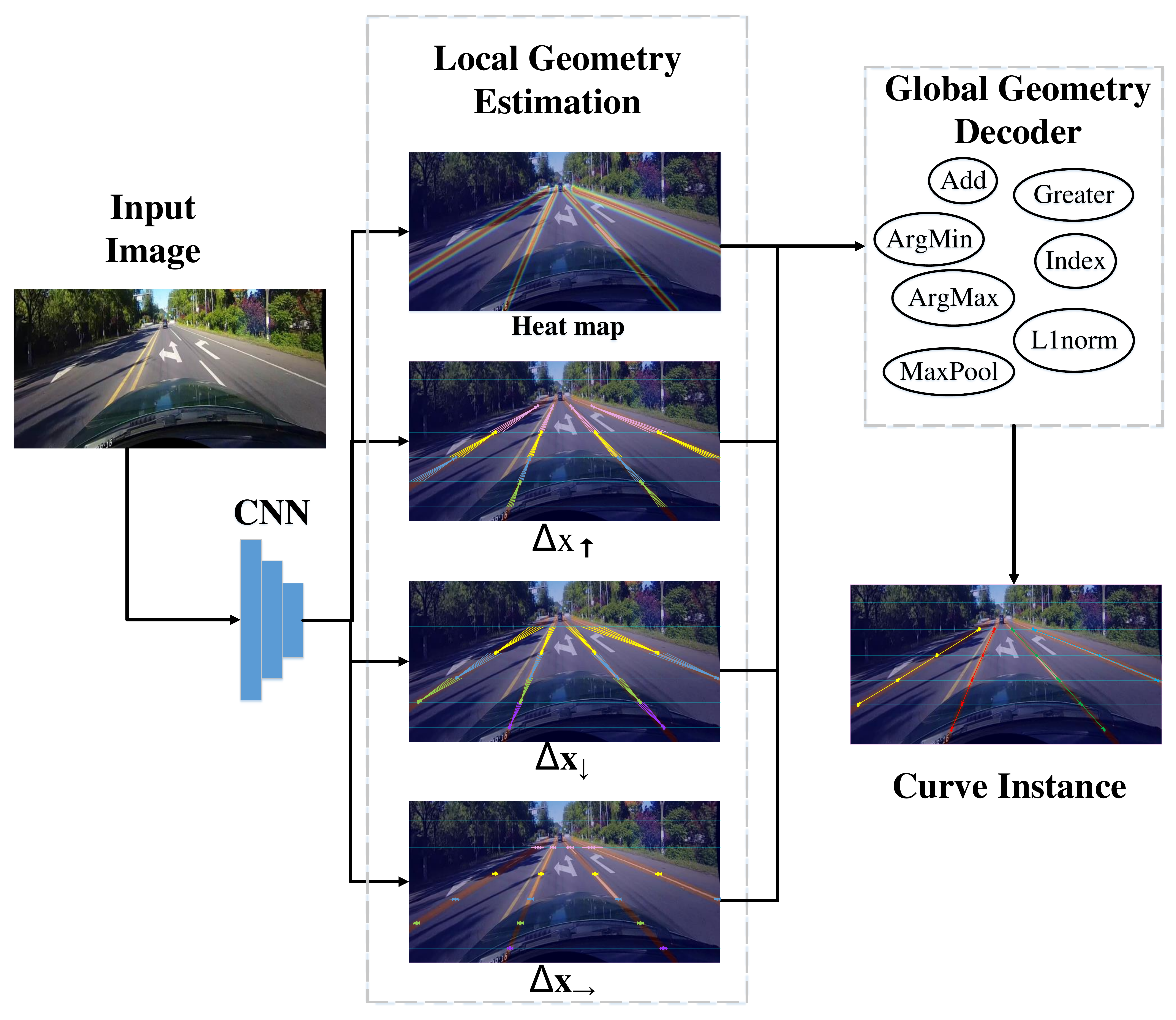}
	\end{center}
	\caption{The inference process of FOLOLane. The network produces 4 logits expressing the geometry of the local curve. The decoder module constituted with low-level operators integrates the local information into curve instances. }
	\label{fig:network}
\end{figure}

\textbf{Key point estimation.}  Motivated by a curve constituted of points, we adopt a keypoint-estimation-based framework. The network firstly outputs a heatmap with the same resolution as input, which models the probability that pixel is a keypoint of the curve. In the training phase, the points set as annotation of the $j-th$ curve are interpolated to be continuous in pixel space as $l_j$. Each pixel of the curve $l_j$ is considered as a key point and yields ground-truth value for neighbors via unnormalized Gaussian kernel. The standard deviation $\sigma_h$ depends on the scale of input, and if the ground-truth value of some pixel is assigned by multiple keypoints, the maximum will be kept.

To deal with the class imbalance problem coming with the sparsity of key points, we employ penalty-reduced focal loss for this head as in ~\cite{law2018cornernet,zhou2019object}, where only pixels with ground truth equal to 1 are considered positive and all others are negative. The penalty from negative pixels arises with the distance to positive, which helps to reduce the influence of ambiguity. We denote the output of $i-th$ pixel at heatmap as $s_i$ and the ground-truth value assigned by Gaussian Kernel as $g_i$. Define penalty coefficients $\hat{g}_i$ and $\hat{s}_i $ as:
\begin{equation}
	\hat{g}_i=
	\begin{cases}
		0 \; &if\; g_i=1 \\
		g_i \; &otherwise 
	\end{cases},\;
	\hat{s}_i=
	\begin{cases}
		s_i \; &if\; g_i=1 \\
		1-s_i \; &otherwise 
	\end{cases},
	\label{eq:reduction_factor}
\end{equation}
and the loss function for heatmap head is constructed as:
\begin{equation}
	Loss_{h}=-\frac{1}{N}\sum^N_i(1-\hat{g}_i)^\beta(1-\hat{s}_i)^\gamma log(\hat{s}_i),
	\label{eq:loss_cls}
\end{equation}
\noindent where $\beta$ and $\gamma$ are tunable hyperparameters, controlling the penalty reduction for ambiguous and simple samples respectively. $N$ is the number of key points in the current image.

Compared with segmentation-based methods, the loss function Eq.\ref{eq:loss_cls} guides the network to learn positive and negative samples of keypoint with reduced supervision from the total pixels, prompting pixels best suited for expressing geometry to the response. An example of the heatmap can be found in Fig.\ref{fig:network} as the first output of the network, the center of lane marker responses highest, and the neighborhood became colder gradually, which helps prevent the noise and redundancy from propagating to subsequent procedures as well.

\textbf{Local geometry construction.} For precise geometry, the second head of the network regresses a vector $[\Delta{x_{\uparrow}}(p), \Delta{x_{\rightarrow}}(p), \Delta{x_{\downarrow}}(p)]^T$, describing the local geometry of the closest curve to pixel $p$. The elements indicate the horizontal offsets to 3 neighboring key points with fixed vertical interval $\Delta y$, which have been colorized for visualization in Fig.\ref{fig:network}. Given the vector, we can simply recover the local curve related to pixel $p$:
\begin{equation}
	\bm{\hat{l}}(p)=
	\left[
	\begin{array}{ccc} 
		\bm{\hat{p}}_{\uparrow}(p) \\
		\bm{\hat{p}}_{\rightarrow}(p) \\
		\bm{\hat{p}}_{\downarrow}(p) \\
	\end{array}
	\right]
	= 
	p+
	\left[
	\begin{array}{ccc} 
		\Delta{x_{\uparrow}}(p)   & -\Delta y \\
		\Delta{x_{\rightarrow}}(p) & 0 \\
		\Delta{x_{\downarrow}}(p) & \Delta y \\
	\end{array}
	\right],
	\label{eq:local_curve}
\end{equation}

\noindent where $\bm{\hat{p}}{\uparrow}(p)$, $\bm{\hat{p}}{\rightarrow}(p)$ and $\bm{\hat{p}}{\downarrow}(p)$ denote the actual location with fixed vertical interval $\Delta y$ to pixel $p$, respectively.

In the training phase, all pixels within a fixed distance from key points of the curve $l$, $N_{\sigma_{g}(l)}$, are taken to compute loss for $\Delta{x_{\uparrow}}, \Delta{x_{\downarrow}}$.

\begin{equation}
	\begin{aligned}
		\bm{Loss_{\uparrow}}(l)=
		\frac{1}{|N_{\sigma_{g}}(l)|}&\Sigma_{p\in{N_{\sigma_{g}}(l)}}
		||\bm{\hat{p}}_{\uparrow}(p) - \varphi(l,f_{y}(p)-\Delta y)||_{1},\\
		\bm{Loss_{\downarrow}}(l)=
		\frac{1}{|N_{\sigma_{g}}(l)|}&\Sigma_{p\in{N_{\sigma_{g}}(l)}}
		||\bm{\hat{p}}_{\downarrow}(p)-  \varphi(l,f_{y}(p)+\Delta y)||_{1},
	\end{aligned}
	\label{eq:loss_updown}
\end{equation}

\noindent where $f_{y}(\cdot)$ denotes the function retrieving vertical coordinate of the pixel, $\varphi(l,y)$ is function retrieving horizontal coordinate of curve $l$ on specific row $y$.

For $\Delta{x_{\rightarrow}}$, a coarse-to-fine strategy is employed:

\begin{equation}
	\begin{aligned}
		\bm{Loss_{\rightarrow}}(l)= & \frac{1}{2|N_{\sigma_{g}}(l)|}\Sigma_{p\in{N_{\sigma_{g}}(l)}}( \\
		&||\bm{\hat{p}}_{\rightarrow}((\bm{\hat{p}}_{\uparrow}(p))) -  \varphi(l,f_{y}(p)-\Delta y)||_{1}  + \\
		&||\bm{\hat{p}}_{\rightarrow}((\bm{\hat{p}}_{\downarrow}(p))) -  \varphi(l,f_{y}(p)+\Delta y)||_{1}),
	\end{aligned}
	\label{eq:loss_same}
\end{equation}

\noindent where the training pixels come from the decoded prediction of $\Delta{x_{\uparrow}}$ and $\Delta{x_{\downarrow}}$ in Eq.\ref{eq:loss_updown}, which is used to compensate for the error in predicting $\Delta{x_{\uparrow}}$ and $\Delta{x_{\downarrow}}$ and keeps in line with the coarse-to-fine behavior in the decoding stage. L1 loss is employed for all the regression terms.

\textbf{Network architecture.} To justify the effectiveness of focusing on local geometry, we adopt light-weight architecture ERFNet~\cite{romera2017erfnet} and BiSeNet~\cite{yu2018bisenet}, which were originally designed for semantic segmentation on mobile devices. During the feature extraction, the encoder abstracts image into downsampled feature map, then the decoder broadcasts the high-level semantics to the same resolution as input. All 4 logits are yielded by the last block of the decoder for saving memory. Most experiments in this paper are performed basing on ERFNet. Since the method is designed for working in real traffic scenarios, which is required to handle the case of a merged or split marker and any number of instances, there is no extra branch specialized for predefined lane markers as in ~\cite{pan2017spatial,hou2019learning}. The final cost function is formulated as
\begin{equation}
	Loss = Loss_{h} + \lambda(Loss_{\uparrow}+Loss_{\downarrow}+Loss_{\rightarrow}),
	\label{eq:final_loss}
\end{equation}

\thispagestyle{empty}
\subsection{Decoder for global geometry}
\label{subsection:decoder}
In the above section, \textit{CNN} produces pixel-wise
heatmap and offset for keypoints in local scope. These local information 
are subsequently integrated into prediction of
global curve. Specifically, the heatmap is used to determine
emergence and termination of curve. The offsets is used
to associate keypoints on same curve instance and refine
geometry further. To this end, we propose two novel and simple algorithms
for decoding the output of \textit{CNN} under different demand scenarios, which responds to
preferences for accuracy and efficiency respectively.


\begin{figure*}[t]
	\begin{center}
		\includegraphics[width=.98\linewidth]{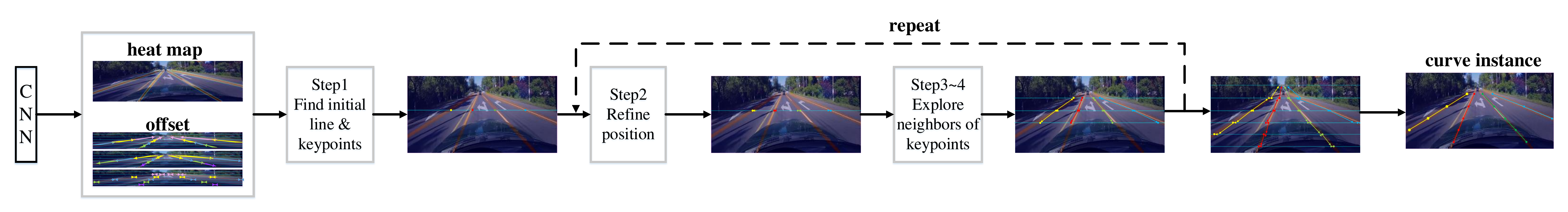}
	\end{center}
	\caption{Illustration of greedy decoding process. All the keypoints found in process have been shown in color. The colored arrows indicate the refinement of position of keypoints, or the prediction of neighboring points. The invalid points is displayed in gray. Finally, the decoded curve instance is represented as set of keypoints in same color. It's best to zoom in the figure and view it in color.}
	\label{fig:decoder_greedy}
\end{figure*}

\begin{figure*}[t]
	\begin{center}
		\includegraphics[width=.98\linewidth]{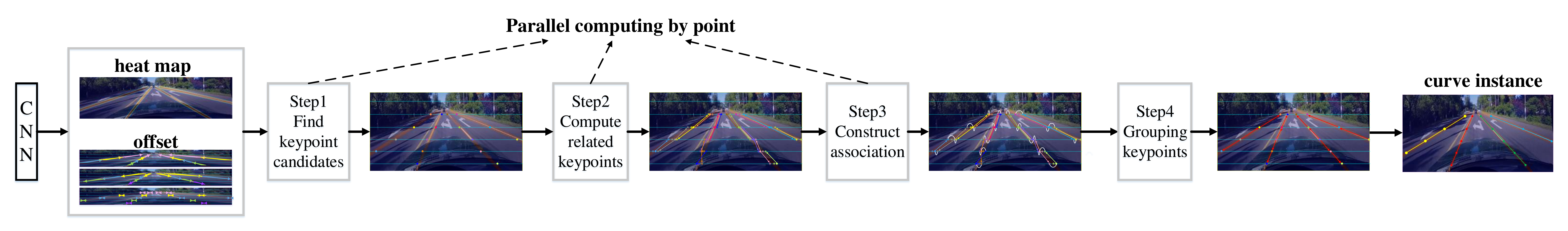}
	\end{center}
	\caption{Illustration of efficient decoding process. Different from greedy decoder, which searches keypoints in an iterative manner, the efficient decoder found all the keypoint candidates at the beginning. For these candidates, the position refinement, neighbor prediction and association construction are perform in one step through parallel computing. The white curve indicates association relationship among keypoints.}
	\label{fig:decoder_efficient}
\end{figure*}

\textbf{Greedy decoder} works through iteratively extending the neighbors of keypoint in a greedy search-like manner. For each input image,
\begin{itemize}[itemindent=1em]
    \item [\textsl{Step1}]
        \textsl{Find the row containing greatest number of local maximum response on heatmap. This row and the points are taken as starting line and \textbf{current} keypoints.}
    
    \item [\textsl{Step2}]
        \textsl{Refine the position of \textbf{current} keypoints. For point $p$, refinement can be formulated as $\hat{p}=p+[\Delta{x_{\rightarrow}}(p),0]^T$}.
    
    \item [\textsl{Step3}]
        \textsl{Explore the vertical neighbors of \textbf{current} points, the coordinates of which can be computed as $p_{\uparrow}=\hat{p}+[\Delta{x_{\uparrow}}(p),-\Delta{y}]^T$ and $p_{\downarrow}=\hat{p}+[\Delta{x_{\downarrow}}(p),\Delta{y}]^T$. }
    
    \item [\textsl{Step4}]
        \textsl{Examine the heatmap value of $p_{\uparrow}$ and $p_{\downarrow}$. If the value reaches threshold $\theta_h$, the corresponding neighboring points is used to update \textbf{current} keypoint, and \textbf{Step2\textasciitilde4} are repeated. Otherwise the search is terminated, all the points searched from one single point are taken as one global curve.}
    
\end{itemize}
To sum up, the decoding algorithm gradually extends the global curve by exploring neighbors of keypoint, and refine the geometry of curve in a coarse-to-fine manner. This algorithm can produce precise geometry of curve, but its low efficiency limits the useability in practical application.
The process have been shown in color in Fig.\ref{fig:decoder_efficient}.


\textbf{Efficient decoder} is proposed in order to solve the inefficiency problem of greedy decoders, which utilizes the parallelism of computing devices. For each image,
\begin{itemize}[itemindent=1em]
    \item [\textsl{Step1}]
        \textsl{Extract rows at equal interval $\Delta y$ on heatmap. On each row, take the points with local maximum response as \textbf{current} keypoints.}
    
    \item [\textsl{Step2}]
        \textsl{For each keypoint $p$, compute three related points as $p_{\rightarrow}=p+[\Delta{x_{\rightarrow}}(p),0]^T$,  $p_{\uparrow}=p+[\Delta{x_{\uparrow}}(p),-\Delta{y}]^T$ and $p_{\downarrow}=p+[\Delta{x_{\downarrow}}(p),\Delta{y}]^T$. }
    
    \item [\textsl{Step3}]
        \textsl{Construct association among \textbf{current} keypoints located in neighboring rows. For a point $p$ in $i$-th row, two points in $(i-\Delta{y})$-th row and $(i+\Delta{y})$-th row will be associated with it, which are closest to the position of $p_{\uparrow}$ and $p_{\downarrow}$ respectively.}
    
    \item [\textsl{Step4}]
        \textsl{Starting with the row with maximum number of current keypoints. According to the association relationship created in Step3, for each current keypoint, all the keypoints associated with it in above/below rows are iteratively taken out as a single group. Each keypoint group is considered as a global curve, and $p_{\rightarrow}$ of points are used to refine geometry of curve further.}
\end{itemize}
The efficient decoding algorithm leverages the parallel computing power of device, to create association among keypoints and refine their position, from step1 to step3. Step4 involves only index operations, thus the time overhead is very low. The process have been shown in color in Fig.\ref{fig:decoder_greedy}.

\thispagestyle{empty}
\section{Experiments}
In this section, firstly we describe the implementation details and evaluation datasets. Followed by the results of comparison with the state-of-the-art, including quantitative and qualitative results. Finally, the discussion of ablation study and generalization are detailed.

\begin{table*}[htbp]
	\begin{center}
		\begin{tabular}{|l|c|c|c|c|c|c|c|}
			\hline
			Dataset & \# Frame & Train & Validation & Test & Resolution & Road type & \# Lane \\
			\hline\hline
			TuSimple & 6408 & 3268 & 358 & 2782 & 1280$\times$720 & highway & $<=$5 \\
			CULane & 133235 & 88880 & 9675 & 34680 & 1640$\times$590 & urban, rural and highway & $<=$4 \\
			\hline
		\end{tabular}
	\end{center}
	\caption{Basic information of two lane marker detection datasets.}
	\label{basic information of two lane marker detection datasets}
\end{table*}

\begin{table*}[htbp]
	\begin{center}
		\setlength{\tabcolsep}{0.2mm}{
			\begin{tabular}{|l|c|c|c|c|c|c|c|c|}
				\hline
				Category & Proportion & SCNN\cite{pan2017spatial} & ENet-SAD\cite{hou2019learning} & ERFNet-E2E\cite{yoo2020end} & SIM-CycleGAN & UFNet\cite{qin2020ultra} & PINet(4H)\cite{ko2020key} & \textbf{FOLOLane} \\
				& & & & & +ERFNet\cite{liu2020lane} & & &\textbf{(ours)} \\
				\hline\hline
				Normal & 27.7\% & 90.6 & 90.1 & 91.0 & 91.8 & 90.7 & 90.3 & \textbf{92.7} \\
				Crowded & 23.4\% & 69.7 & 68.8 & 73.1 & 71.8 & 70.2 & 72.3 & \textbf{77.8} \\
				Night & 20.3\% & 66.1 & 66.0 & 67.9 & 69.4 & 66.7 & 67.7 & \textbf{74.5} \\
				No line & 11.7\% & 43.4 & 41.6 & 46.6 & 46.1 & 44.4 & 49.8 & \textbf{52.1} \\
				Shadow & 2.7\% & 66.9 & 65.9 & 74.1 & 76.2 & 69.3 & 68.4 & \textbf{79.3} \\
				Arrow & 2.6\% & 84.1 & 84.0 & 85.8 & 87.8 & 85.7 & 83.7 & \textbf{89.0} \\
				Dazzle light & 1.4\% & 58.5 & 60.2 & 64.5 & 66.4 & 59.5 & 66.3 & \textbf{75.2} \\
				Curve & 1.2\% & 64.4 & 65.7 & \textbf{71.9} & 67.1 & 69.5 & 65.6 & 69.4 \\
				Crossroad & 9.0\% & 1990 & 1998 & 2022 & 2346 & 2037 & \textbf{1427} & 1569 \\
				\hline
				Total & - & 71.6 & 70.8 & 74.0 & 73.9 & 74.4 & 72.3 & \textbf{78.8} \\
				\hline
			\end{tabular}}
	\end{center}
	\caption{Performance of different methods on CULane testing set, with IoU threshold=0.5. For crossroad, only FP are shown.}
	\label{culane results}
\end{table*}

\begin{table}[htbp]
	\begin{center}
		\setlength{\tabcolsep}{1.3mm}{
			\begin{tabular}{|l|c|c|c|}
				\hline
				Method & Accuracy(\%) & FP & FN \\
				\hline\hline
				SCNN\cite{pan2017spatial} & 96.53 & 0.0617 & \textbf{0.0180} \\
				LaneNet(+H-Net)\cite{neven2018towards} & 96.40 & 0.0780 & 0.0244 \\
				EL-GAN\cite{ghafoorian2018gan} & 96.39 & 0.0412 & 0.0336 \\
				PointLaneNet\cite{chen2019pointlanenet} & 96.34 & 0.0467 & 0.0518 \\
				FastDraw\cite{philion2019fastdraw} & 95.2 & 0.0760 & 0.0450 \\
				ENet-SAD\cite{hou2019learning} & 96.64 & 0.0602 & 0.0205 \\
				ERFNet-E2E\cite{yoo2020end} & 96.02 & 0.0321 & 0.0428 \\
				PINet(4H)\cite{ko2020key} & 96.75 & \textbf{0.0310} & 0.0250 \\
				\hline
				\textbf{FOLOLane(ours)} & \textbf{96.92} & 0.0447 & 0.0228 \\
				\hline
			\end{tabular}}
	\end{center}
	\caption{Performance of different methods on TuSimple testing set.}
	\label{tusimple results}
\end{table}

\subsection{Implementation Details}
\label{subsection:implementation details}
We first resized the width of an image to 976 and kept the aspect ratio on both datasets. The $\Delta{y}$ was set as 10 pixels for a trade-off between precision and efficiency. The weight $\lambda$ for loss function in Eq.[\ref{eq:final_loss}] was set as 0.02. For optimization, we used Adam optimizer and poly learning rate schedule with an initial learning rate of 0.001. Each mini-batch contained 16 images per GPU and we trained the model using 8 V-100 GPUs for 40 epochs on CULane and 200 epochs on TuSimple, respectively. To reduce overfitting, we used a 0.3 probability of dropout and weight decay with 0.0001. Furthermore, we also applied data augmentation, including random scaling, cropping, horizontal flipping, random rotation, and color jittering, which have been proved to be effective. In the testing phase, we set the threshold of lane existence confidence as 0.5.

As illustrated in Table \ref{basic information of two lane marker detection datasets}, the basic information of TuSimple and CULane datasets are detailed. And for evaluation criteria, we follow the official metric used in \cite{tusimple} and \cite{pan2017spatial}.

\thispagestyle{empty}
\subsection{Results}

\begin{figure*}[htbp]
	\begin{center}
		\includegraphics[width=.9\linewidth]{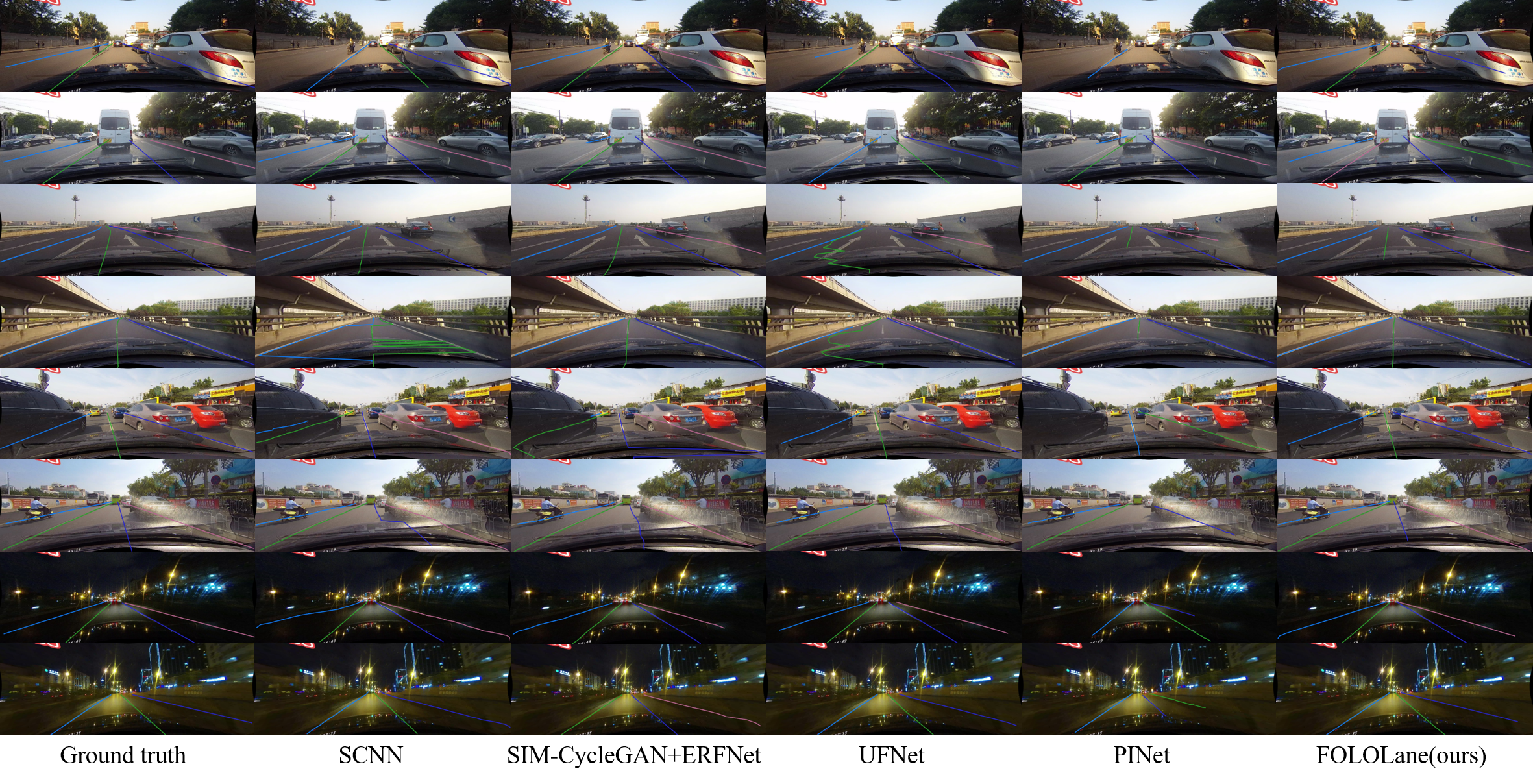}
	\end{center}
	\caption{Visualized results of SCNN, SIM-CycleGAN+ERFNet, UFNet, PINet and FOLOLane on CULane testing set.}
	\label{fig:qualitative results}
\end{figure*}

In this section, we show the results on two lane detection datasets. In all experiments, ERFNet\cite{romera2017erfnet} is used as our baseline network if not specially mentioned.

\textbf{Quantitative results.} To verify the effectiveness of our proposed method, we compared it with state-of-the-art algorithms based on either segmentation or object detection, including SCNN\cite{pan2017spatial}, LaneNet(+H-Net)\cite{neven2018towards}, EL-GAN\cite{ghafoorian2018gan},  PointLaneNet\cite{chen2019pointlanenet}, FastDraw\cite{philion2019fastdraw}, ENet-SAD\cite{hou2019learning}, ERFNet-E2E\cite{yoo2020end}, SIM-CycleGAN+ERFNet\cite{liu2020lane}, UFNet\cite{qin2020ultra} and PINet\cite{ko2020key}.

As illustrated in Table \ref{culane results}, the proposed method achieves a new SOTA result on the CULane testing set with a 78.8 F1 measure. Compared with the best model as far as we know, PINet(4H), our method outperforms almost all of the scenarios, whose F1 measure improves 4.4\%. Because of local occlusions and fogged traffic lines, PINet shows degraded performance in some categories, such as Crowded, Arrow and Curve. Although our method and PINet are both based on key points estimation, in the aforementioned categories, our method outperforms PINet with 5.5\%, 5.3\%, and 3.8\% F1 measure improvements respectively, which indicates our local geometry modeling model and bottom-up pipeline have better lane marker representation capabilities. Besides, an interesting point is that SIM-CycleGAN+ERFNet, which aims at dealing with low light conditions using CycleGAN, is not comparable to our lane marker detection model in the night and dazzle light scenarios, which implies that our approach is of better generalization ability even than GAN augmented data.

The results of different methods on the TuSimple testing set are shown in Table \ref{tusimple results}. Due to the limited scale (train/test:3.3k/2.8k) and homogeneous scenario (highway), most methods achieved near-saturated accuracy (more than 96\%). Despite this, our method still outperforms the 2nd by 0.17\%, close to the difference between 2nd and 4th.

\textbf{Qualitative results.} We also show qualitative results of the proposed method and SCNN, SIM-CycleGAN+ERFNet, UFNet, PINet on the CULane testing set. As shown in Fig.\ref{fig:qualitative results}, our method focusing on local geometry and bottom-up strategy helps to distinguish the occlusion of crowded roads and the missing lane marker clues. Through keypoint estimation, the proposed method could yield a smoother and more accurate curve than the others do. Even though in night and dazzle light scenarios, the predicted results are still satisfactory. In conclusion, the proposed method leads to visible improvements in lane marker detection among recently developed segmentation-based and regression-based approaches.

\thispagestyle{empty}
\subsection{Ablation Study}
To investigate the effects of the locally based designs, an ablation study is carried out on the CULane dataset. The experiments are all conducted with the same settings as described in Sec.~\ref{subsection:implementation details} if not specially mentioned.

\begin{table}[htbp]
	\begin{center}
		\setlength{\tabcolsep}{0.5mm}{
			\begin{tabular}{|c|c|c|c|c|c|c|c|c|c|}
				\hline				
				\multicolumn{2}{|c|}{Heatmap} &
				\multicolumn{2}{c|}{Coarse-to-fine} &
				\multicolumn{2}{c|}{Decoder} &
				\multicolumn{2}{c|}{Architecture} &
				\multicolumn{1}{c|}{\multirow{1}{*}{F1}} & 
				\multicolumn{1}{c|}{\multirow{1}{*}{Rt.}} \\
				\cline{1-8}
				Se. & Ke. & @ test & @ train & Gre. & Eff. & ERF & BiSe & \multicolumn{1}{c|}{} & \multicolumn{1}{c|}{} \\
				\hline\hline
				\checkmark &  &  &  & \checkmark &  & \checkmark & & 74.2 & - \\
				 & \checkmark &  &  & \checkmark &  & \checkmark & & 76.6 & - \\
				 & \checkmark & \checkmark &  & \checkmark &  & \checkmark & & 77.5 & - \\
				 & \checkmark & \checkmark & \checkmark & \checkmark &  & \checkmark & & \textbf{78.8} & 25ms \\
				 & \checkmark & \checkmark & \checkmark &  & \checkmark & \checkmark & & 78.3 & 16ms \\
				 & \checkmark & \checkmark & \checkmark &  & \checkmark & & \checkmark & 77.5 & \textbf{9ms} \\
				\hline
		\end{tabular}}
	\end{center}
	\caption{Ablation studies on CULane testing set. \textbf{Se.}: Semantic segmentation. \textbf{Ke.}: Keypoint estimation based heatmap. \textbf{@test}: use $\Delta{x_{\rightarrow}}$ to refine the geometry of the curve in testing. \textbf{@train}: use coarse-to-fine strategy to sample training data for $\Delta{x_{\rightarrow}}$ in training. \textbf{Gre.}: Greedy decoding. \textbf{Eff.}: Efficient decoding. \textbf{Rt.}: runtime.}
	\label{ablation studies}
\vspace{-1.em}
\end{table}

\begin{figure*}[htbp]
	\begin{center}
		\includegraphics[width=.75\linewidth]{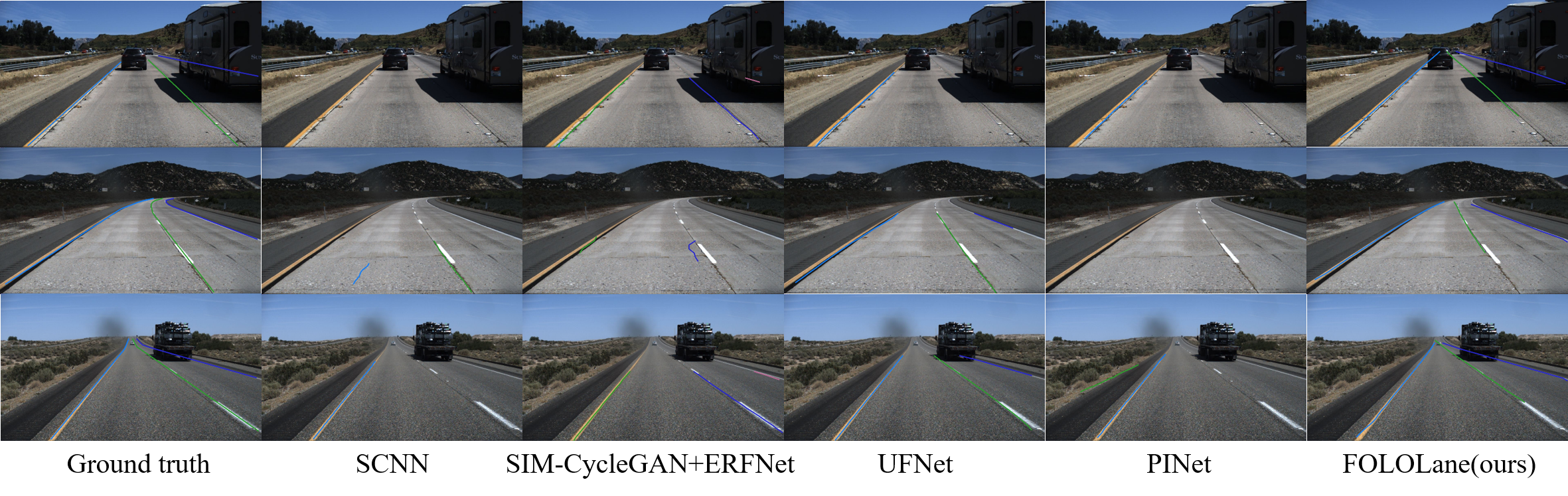}
	\end{center}
	\caption{Visualization results of generalizing SCNN, SIM-CycleGAN+ERFNet, UFNet, PINet and FOLOLane on TuSimple testing set.}
	\label{fig:generalization results}
\end{figure*}

\textbf{Key point estimation.} Different from segmentation-based solutions, our key point estimation method focuses on the center of the lane marker, which achieves a impressive result. Table \ref{ablation studies} shows that the proposed method improves the F1 measure from 74.2 to 76.6, which indicates that the suppression of ambiguous and noisy pixels helps achieve accurate geometry and fewer false positives, improving the performance of a system in turn.

\textbf{Coarse-to-fine geometry refinement.} During both network training and instance decoding, we adopt a coarse-to-fine geometry refinement for a more accurate position of key points. In the training phase, the training pixels come from the decoded prediction of $\Delta{x_{\uparrow}}$ and $\Delta{x_{\downarrow}}$. In the inference phase, The predicted $\Delta{x_{\rightarrow}}$ is employed to refine the position of initial key points and newly explored neighboring key points. The results of different configurations are shown in Table \ref{ablation studies}. Only using coarse-to-fine in inference improves the F1-measure 0.9\%. When coarse-to-fine is extended to training, the performance outperforms that of uniform sampling in $N_{\sigma_{g}}(l)$ significantly by 1.3\%. The result shows that the direct prediction leads to suboptimal position estimation and our coarse-to-fine strategy could guide the spatially most related representation to capture the geometry of the curve and achieve a more accurate prediction.

\textbf{Efficiency-oriented implementation.} As mentioned in  Sec.~\ref{subsection:decoder}, efficient decoding is aimed at real-time processing. The main difference from a greedy decoder is that the iteration of decoding neighboring key points is replaced by parallel processing. The parallel decoding significantly improves the efficiency, which achieves 16 ms (64\%) runtime gains than greedy decoder at the cost of 0.8\% performance degradation. The reason can be attributed to the lack of local optimal estimation in each iteration of greedy decoding. 

To maximize efficiency for application, we further replace the basic network from ERFNet to BiSeNet, which is a real-time semantic segmentation network originally designed for mobile devices. Since the output of BiSeNet is 8 times downsampled from the input size, real-time performance is achieved by reaching  more than 100 fps and 77.5 F1 measure simultaneously, which is still the best state-of-the-art results excluding the accuracy-oriented version of our approach. On the other side, the experiment also proves the compatibility of the proposed system, which can be readily adapted for more powerful and efficient network architectures up to date.

\subsection{Generalization}
To further verify the generalization of our proposed method, we employ the checkpoint trained from the CULane training set to inference on the TuSimple testing set. To our knowledge, this is the first attempt to investigate the generalization between these two widely used datasets. Table \ref{generalization of different methods from CULane training set to TuSimple testing set} shows that the proposed method achieves obvious superiority with an accuracy of 84.36\%, which surpasses other methods by a significant margin of nearly 20\%. The SCNN and PINet(4H) approaches suffer most from the generalization ability, which decreases 90\% and 60\% respectively. The generalized visualization results on the TuSimple testing set are shown in Fig.\ref{fig:generalization results}. This result indicates that the simplified task and the compact output of the network reduce the demand for model capacity and training data, the resulting stableness and efficiency in training finally lead to advantageous generalization to other domains, which shows promising potential for application.

\begin{table}[htbp]
	\begin{center}
		\setlength{\tabcolsep}{0.2mm}{
			\begin{tabular}{|l|c|c|c|}
				\hline
				Method & Accuracy(\%) & FP & FN \\
				\hline\hline
				SCNN\cite{pan2017spatial} & 0.29 & \textbf{0.0068} & 1.0 \\
				SIM-CycleGAN+ERFNet\cite{liu2020lane} & 62.58 & 0.9886 & 0.9909 \\
				UFNet\cite{qin2020ultra} & 65.53 & 0.5680 & 0.6546 \\
				PINet(4H)\cite{ko2020key} & 36.31 & 0.4886 & 0.8988 \\
				\hline
				\textbf{FOLOLane(ours)} & \textbf{84.36} & 0.3964 & \textbf{0.3841} \\
				\hline
		\end{tabular}}
	\end{center}
	\caption{Evaluation of generalization ability of different methods from CULane training set to TuSimple testing set.}
	\label{generalization of different methods from CULane training set to TuSimple testing set}
\vspace{-1.em}
\end{table}

\thispagestyle{empty}
\section{Conclusion and Future Work}
In this paper, we propose a local-based bottom-up solution for lane detection. Experimental results show the keypoint estimation and the coarse-to-fine refinement strategy circumvent the influence from ambiguous and noisy pixels, effectively improves the accuracy of curve geometry. More importantly, the principle of focusing on local geometry and the bottom-up pipeline have been proved to be particularly resultful, which significantly simplifies the task by reducing the dimension of the output of CNN and is believed to be the principal cause of the excellent performance and generalization capacity.

The proposed method also shows superiority in adaptation to the rapid evolution of neural networks for performance and efficiency. We have plan to incorporate more powerful architectures into \textbf{FOLOLane} framework, e.g. the ones with self-attention mechanism, to improve the performance further. We also want to use \textbf{FOLOLane} on MindSpore\footnote{https://www.mindspore.cn/}, which is a new deep learning computing framework. These problems are left for future work.


{\small
\bibliographystyle{ieee_fullname}
\bibliography{egbib}
\thispagestyle{empty}
}

\end{document}